%% file: distr-estimation-control.tex
\documentclass[letter]{article}
\usepackage[english,portuguese]{babel}
\usepackage[T1]{fontenc}
\usepackage[utf8]{inputenc}

\usepackage{ulem} 
\usepackage{latexsym}

\usepackage{amsmath, amsthm, amssymb, amsfonts, amsbsy}
\usepackage{color,soul}

\usepackage{multirow}
\usepackage{colortbl}
\usepackage{supertabular}
\usepackage{pdflscape}

\usepackage{graphics}
\usepackage{graphicx}
\usepackage{epsfig}
\usepackage{subcaption}
\usepackage{dcolumn}
\usepackage{bm}
\usepackage{booktabs}
\usepackage{rotating}
\usepackage{multirow}

\providecommand{\keywords}[1]
{
  \small	
  \textbf{\textit{Keywords---}} #1
}

\usepackage[plainpages=false]{hyperref}
\hypersetup{
             colorlinks=false,
             citecolor=red,
             breaklinks=true,
             bookmarksnumbered=true,
             bookmarksopen=true,
             pdftitle={},
             pdfauthor={Vasques,Soares,Gomes},
             pdfsubject={Active online localization},
             pdfcreator={Authors},
             pdfkeywords={Active learning, distributed learning,
               distributed control, target localization, Fisher
               Information Matrix, Consensus + Innovations}}
\usepackage{float}

\newcommand{\minimize}{\operatornamewithlimits{minimize}}

\newtheorem{theorem}{Theorem} 

\newtheorem{proposition}[theorem]{Proposition}
\newtheorem{remark}[theorem]{Remark}

\title{ISEE.U: Distributed online active
  target localization with unpredictable targets}

\author{Miguel
  Ara\'{u}jo Vasques,  Cl\'audia Soares*, and Jo\~ao Gomes}

\begin{document}

\selectlanguage{english}
\maketitle

\begin{abstract}
\input{03.abstract}

\end{abstract}


\section{Introduction}
\label{sec:intro_prob}


Real-world applications such as logistics, security, minerals and oil
exploration, personal and vehicle navigation, wireless communications,
and surveillance, just to mention a few, struggle for achieving a
solution for medium-high accuracy localization of some non cooperative
targets.


Most approaches for
range-based localization do not assume agents can control the network
motion to improve localization accuracy. Thus, a passive localization
algorithm solely relies on a stream of sensor data. One of the first
approaches for active localization is~\cite{fox1998}, where one robot
attempts to self-localize with a Markovian approach: computing a
belief for a discretized map of the region of interest, given sensor
measurements, and maximizing the entropy of its next movement.


However, when we envision large teams of moving artificial agents,
with high-level tasks, like intercepting an intruder, the
computational paradigm should accommodate scalability concerns. Having
scalability in mind, we define a distributed algorithm as a procedure
running at all agents, with similar computations at each one,
demanding information exchanges only with one-hop neighbors. We aim to
develop a scalable, distributed method which is computationally
simple, fast, and robust to changes in the network configuration, like
broken communication links and exiting and entering nodes in the
network. Further, we do not rely on a dynamical model of the
uncooperative moving targets, because they can be adversarial.


Further, real-world 
teams of agents have to deal with unaccounted changes in the environment and
constantly moving targets. Such scenarios demand fast response,
lightweight algorithms that return online estimates, fast enough for
the operation to achieve its goal.

\paragraph*{Related work}
\label{sec:related-work}

Target localization is a long-standing problem, and with many
different setups, assumptions, and data models considered in prior
work. Target localization can be seen as a variant of the network localization problem~\cite{beck2008,stoica2006,8493177,SOARES2021108066,domingos2022robust}. One line of work in the target localization problem considers static sensors as~\cite{Taylor2006},
where the problem is formulated in a Bayesian framework, or
in~\cite{EkimGomesXavierOliveira2009}, where the proposed estimator
arises from an optimization framework. Another take on the same
problem is the optimal fixed sensor placement for target tracking,
pursued in~\cite{Ucinski2004} and, more recently,
in~\cite{Xu2017}. Mobile sensor networks add to the static setting in
terms of area coverage, adaptability to the environment, a target
behavior, and, of course, improve the quality of sensing and
estimation.  Cooperative active sensing harnesses mobility and
cooperation in teams of agents to boost performance and flexibility in
target localization and tracking~\cite{Zhou2011,Zhou2017,Renfrew2018}.
The work of Hook et al.~\cite{Hook2015} develops an active
localization solution using bearing information. The centralized
algorithm has an online and offline flavor and assumes static
targets. In our work, in contrast, we assume range data, with a fully
distributed online operation and moving targets.

Most of the methods are based on linear Kalman
filtering~\cite{olfati2012,rad2011,liu2014,morbidi2013}, and in the
Extended Kalman Filter (EKF)~\cite{martinez2006,wei2014}, while other
solutions are based on Bayesian
estimation~\cite{tisdale2009,banfi2015,esmailifar2015}. In~\cite{olfati2012},
the authors present an active localization algorithm, assuming that
the information quality of a sensor increases whenever the distance to
the sensor decreases, but the data model prescribes noisy linear
measurements of the target \emph{position}. The present work also
considers that measurement noise depends on the distance to the
measured target, but we do not assume we can linearly sense
position. Instead, our data model requires only noisy range
measurements, which are a nonlinear one-dimensional random function of
the true 2D or 3D position. This data model comes as a practical
applicational need, because there is a broader choice on inexpensive
ways to acquire range measurements than ways to measure a linear
function on position. Rad et al.~\cite{rad2011} approaches target
localization from range measurements by linearizing the measurement
model and applying a Kalman Filter to the linearized problem. Their
model focuses only on position estimation and does not consider active
localization. In~\cite{liu2014} a Kalman Filter is used in combination
with a Particle Filter, again assuming pose measurements, with a
linear measurement model. Also, using noisy linear measurements of the
position, the interesting work in~\cite{morbidi2013} presents a Kalman
Filtering approach, where noise is modelled as a quadratic function of
the distance to the target. The EKF is proposed in~\cite{martinez2006}
to deal with nonlinear measurements, namely with distances, in a
context where the sensing agents are allowed to move on the boundary
of a convex set, whose interior must contain the target. Another
approach using the EKF is~\cite{wei2014}, where the data model also
includes proprioceptive sensors of the agents.  A few papers dealt
with active decentralized localization based on bearings and linear
velocities of agents, like~\cite{spica2016active}. Our work assumes
knowledge of ranges only and no information on velocity. Further, we
are not interested in estimating private parameters at each agent
(self-localization), but to produce a control that will allow for
enhanced and network-wide distributed estimation of a single
parameter: the location of a common target.  A very recent
paper~\cite{salaris2019online} also considers online self-localization
of individual agents, now with two range measurements to fixed
landmarks, using an EKF. It is, thus, different from our present work
in both goals and approach.

Among the Bayesian filter methods, we point to the work
in~\cite{tisdale2009}, where the authors explore a greedy technique to
maximize information gain in path planning for target search and
localization, but the scheme depends on including the sensor model in
the control, thus relying on proprioceptive measurements. The work
in~\cite{banfi2015} focuses on multi-target tracking and coverage by a
team of robots cooperating in order to maximize tracking
fairness. This approach demands, however, that the motion of the
target be modeled, albeit with some uncertainty. The authors
of~\cite{banfi2015} present an interesting multi-objective
optimization problem where they maximize the average detection rate
over all targets and simultaneously minimize the standard deviation of
this average detection rate, to represent coverage fairness. They opt
for a receding horizon control where the planner is implemented
through an integer linear program to plan robot movements as a
directed graph over the discretized region of operation. The
multi-objective problem is converted to a uni-objective one via linear
combination, depending on a tunable parameter. This approach to target
tracking uses the motion capability of the network to improve the
average detection rate in a way that, albeit covering the fairness
issue, does not consider that proximity to targets might improve
localization accuracy.  Reference~\cite{esmailifar2015} features a
recursive Bayesian estimator with the objective of creating
trajectories to improve the quality of the measurements. To accomplish
this goal, the authors want to drive each vehicle to a point where the
probability of target detection is maximum. To implement the filter
the authors postulate a dynamic model for the target. Such assumption
can hinder the generality of the method, because targets can naturally
or adversarially exhibit large mismatches with such expected
behaviors.

Our work builds upon a deterministic approach, like~\cite{soatti2017},
where the network localization problem is addressed from noisy
Received Signal Strength measurements and a linear model with the
cooperating network positions as unknowns.  Our method assumes each
node can only access range measurements to the target, and an accurate
estimate of its own position. All of the prior art described so far
assumes a less stringent data model. A comparable method --- used as
baseline --- was presented in~\cite{meyer2015}. The referred work is a
distributed method that uses Bayesian statistics to compute a belief
fed to a \textit{Minimum Mean Square Error} (MMSE) estimator. Then, a
particle filter is used to compute a sample-based approximation for
sequential state estimation. The motion of the agents is computed
using an information-seeking controller, which maximizes the negative
posterior joint entropy using a gradient ascent iteration.

\begin{figure}[tb]
  \centering
  \includegraphics[width=0.4\linewidth]{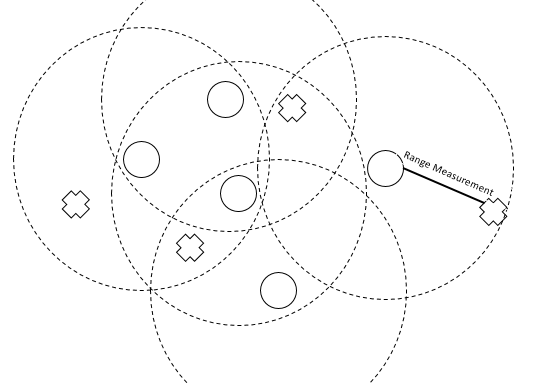}
  \caption[Setup for the considered solution.]{Circles represent
    agents, crosses represent targets and dashed lines are the agents'
    fields-of-view (FoVs). Agents can measure their noisy distance to
    targets that are within their FoV.}
  \label{fig:setup}
\end{figure}

\paragraph*{Overview of the Approach and Contributions}
\label{sec:over}
The setup of our solution comprises (i) a network of mobile agents
with a finite \textit{Field of View} (FoV), where each agent can
measure the distance to one or more targets in the FoV, and (ii)
communication between agents with a finite range. A visual
representation of this setup is shown in Figure \ref{fig:setup}.

We will consider the target localization optimization problem
formulated according to the \textit{Squared-Range-Based Least Squares}
(SR-LS) approach~\cite{beck2008}. We linearize the problem and we
present ISEE.U, a novel, fast, simple to implement, fully distributed
linear algorithm approximating the \textit{Minimum Variance Unbiased}
(MVU) estimator. ISEE.U finds both estimates for the
target positions, and the covariance matrices of those estimates. The
ISEE.U estimator relies on a novel consensus + innovations method
where each agent exchanges information with its one-hop neighbors,
returning accurate covariance matrices, in a distributed manner.

The estimated covariance matrices can be used to design the control
law for the network. From the eigenvalues of the covariance matrix,
which is the inverse of the \textit{Fisher Information Matrix}
(FIM), we can compute the volume of an error ellipsoid for each
target. The goal is to minimize the volume of that error ellipsoid in
order to improve accuracy of the estimated target position: the next
position of the agent will always be the one that minimizes the volume
of the error ellipsoid.

The main results provided in this paper are:
\begin{itemize}
\item A distributed method to both estimate the target positions and
  control of the network agents. This is a simple to implement, fast
  and flexible method demanding only range measurements to targets;
\item The distributed estimates and control are computed through a
  novel, fast consensus + innovations scheme;
\item Top performance is attained when compared to an equivalent
  state-of-the-art algorithm using a particle filter approach.
\end{itemize}
Matlab source code for the method will be made available upon
acceptation.

This paper is structured as follows: in Section~\ref{sec:probform} we
present the problem; in Section~\ref{sec:method} we detail our
proposed method; in Section~\ref{cap:numres} we demonstrate the
performance of our method through simulations; in
Section~\ref{sec:conclusion} we draw conclusions about the work
developed and propose some open avenues for future research.

\section{Problem Formulation}
\label{sec:probform}
We define a network composed by $n(t)$ agents and $K(t)$ targets. The
position of agent $i$ at time $t$ is given by
$\textbf{s}_i(t) \in \mathbb{R}^d$, where $d = 2$ or $d = 3$ is the
dimension of the space considered, and
$\textbf{p}_k(t) \in \mathbb{R}^d$ represents the position of target
$k \in \{1, \dots, K(t)\}$ at discrete time instant $t$.

Consider also the communications graph $G(t) = (V(t),E(t))$,
representing the communication network established between cooperating
agents, where the set of nodes $V(t) = \{1,...,n(t)\}$ corresponds to
the set of mobile agents. The set
$E(t) = \{ i \sim j \: : \: i,j \in V(t)\}$ aggregates all edges. Each
edge represents a communication link between two agents in the
network.

Finally, consider the measurement graph
$\mathcal{G}(t) = (\mathcal{V}(t),\mathcal{E}(t))$, representing the
network of range measurements of the agents relative to the
targets. The nodes of the graph are given by
$\mathcal{V}(t) = V(t) \cup \mathcal{T}(t)$, where
$\mathcal{T}(t) = \{1,...,K(t)\}$ and where each node represents one
agent or target of the network. In this graph the edges are given by
$\mathcal{E}(t) = \{ i \sim k \: : \: i \in V(t),k \in
\mathcal{T}(t)\}$ where each edge is a range measurement between one
agent and one target.  The range measurement obtained by agent $i$ to
target $k$ at time $t$ is
\begin{equation}
\label{eq:defrange}
r_{ik}(t) = \|\textbf{s}_i(t) - \textbf{p}_k(t)\| + w_{ik}(t),
\end{equation}
where $w_{ik}(t) \sim \mathcal{N}(0,\sigma^2)$ is a white Gaussian
noise term with zero mean and standard deviation $\sigma$.

The data available are the positions of the agents $\textbf{s}_i(t)$
and range measurements $r_{ik}(t)$. The method output is an estimate
for the positions of the targets $\textbf{\^p}_k(t)$ and the control
that defines the next positions for the agents $\textbf{s}_i(t+1)$.

For simplicity of the mathematical exposition, we will consider a
setup consisting of only one target and multiple agents in
Section~\ref{sec:method}. To scale this method to more targets it is
only necessary to change the dimensions of the matrices and repeat
their entries.

\section{ISEE.U: distributed active localization for a mobile
  network of agents}
\label{sec:method}
We start by analyzing the problem of finding the position of the
target $\textbf{p}(t)$ that minimizes the SR-LS cost. As discussed in
the previous section, we will consider~$K(t) = 1$ target and we will
simplify the notation for the noisy range
measurement~$r_{ik}(t) = r_{i}(t)$. The problem of finding the target
position~$\textbf{p}$ using the SR-LS formulation is
\begin{equation} \label{eq:lscost}
\minimize_{\textbf{p}(t) \in \mathbb{R}^d} \sum^{n(t)}_{i=1} (\|\textbf{s}_i(t)
- \textbf{p}(t)\|^2 - r_i^2(t))^2. 
\end{equation}
This problem is nonconvex and thus hard to solve. An option
is to convexify the problem, thus providing an accurate approximation
to the SR-LS problem. To streamline the mathematical presentation, we
will drop the time index whenever the supression does not impair
clarity.

To approximate the nonconvex cost in~\eqref{eq:lscost} we manipulated
equation~\eqref{eq:defrange} and obtained
\begin{equation}
\label{eq:linprob}
\textbf{y} = \textbf{A}\textbf{x} + 2\bm{\Xi}\textbf{w}+(\textbf{w}\circ\textbf{w}),
\end{equation}
where vectors~$\mathbf{y}$, $\mathbf{x}$, and $\mathbf{w}$ are defined
as $\textbf{y} = (y_1, \cdots, y_n)$, with $y_i = r_i(t)^2-\|\textbf{s}(t)_i\|^2$, $\textbf{x} = \begin{bmatrix}
\|\textbf{p}(t)\|^2\\
\textbf{p}(t)
\end{bmatrix},$ and $\textbf{w} =(w_1, \cdots, w_n)$
and matrix $\textbf{A}$, whose lines are of the form $a_i^T = (1, \: -2\textbf{s}_1^T(t))$. Matrix $\bm{\Xi}$ is a diagonal matrix with nonzero entries defined as $\bm{\Xi}_{ii} = \|\textbf{s}_i - \textbf{p}\|$.
Finally, $(\textbf{w}\circ\textbf{w})$ is the Hadamard product of the
noise vector $\mathbf{w}$.
The standard deviation of the noise for each agent was defined
considering the influence of distance as
$\sigma_i = \beta \|\textbf{s}_i - \textbf{p}\|\hat{\sigma}$, where
$\hat{\sigma}$ is the default standard deviation for the agents,
$\beta$ is a scalar variable that dictates the influence of
distance on the noise and $\beta\hat{\sigma}$ is considered a
dimensionless quantity.

The data model in~\eqref{eq:linprob} still implies a nonconvex
estimator, which is not fit for real-time distributed setups. As seen
in \cite{beck2008} and \cite{stoica2006}, one possible way to
convexify it is to neglect the dependence of the coordinates of the
unknown variable~$x_1 = \|(x_{2}, \cdots, x_{d})\|$. Doing so will
lead to a linear estimation problem. Considering different statistical
properties of the measurement noise will allow for two formulations
with different characteristics, as described in the next section.

\subsection{Neglecting the squared noise term in~\eqref{eq:linprob}}
\label{sec:unbiased}
We note that the term
$(\textbf{w}\circ\textbf{w})$ in~\eqref{eq:linprob} is much smaller
than the linear noise term multiplied by the distance between the
agents and targets. We neglect the squared noise term and obtain
the following linear model
\begin{equation}
\label{eq:linunb}
\textbf{y} = \textbf{A}\textbf{x} + 2\bm{\Xi}\textbf{w}.
\end{equation}
Now it is possible to apply a simple MVU linear estimator~\cite{kay}
to problem~\eqref{eq:linunb}, given by
\begin{equation}\label{eq:estimator}
\hat{\textbf{x}} =
(\textbf{A}^T\textbf{C}_{\textbf{y}}^{-1}\textbf{A})^{-1}\textbf{A}^T\textbf{C}_{\textbf{y}}^{-1}\textbf{y}
\end{equation}
\begin{equation}\label{eq:covest}
\textbf{C}_{\hat{\textbf{x}}} = (\textbf{A}^T\textbf{C}_{\textbf{y}}^{-1}\textbf{A})^{-1},
\end{equation}
where $\textbf{C}_{\hat{\textbf{x}}}$ is the covariance matrix of the
estimator. The inverse of the covariance matrix of the range
measurements $\textbf{C}_{\textbf{y}}$ is
\begin{align}\label{eq:cyinvunb}
\textbf{C}_{\textbf{y}}^{-1} = \begin{bmatrix}
\frac{1}{4\|\textbf{s}_1-\textbf{p}\|^2\sigma_1^2} & & \multirow{2}{*}{\huge$0$}\\
\multirow{2}{*}{\huge$0$} & \ddots &\\
 & & \frac{1}{4\|\textbf{s}_n-\textbf{p}\|^2\sigma_n^2}.
\end{bmatrix}
\end{align}
Since in the considered approximation~$\textbf{y}$ is a normally
distributed vector and the estimator is just a linear transformation
of it, the statistical performance of~$\hat{\textbf{x}}$ is completely
specified and given by
\begin{equation*} \label{eq:stpropunb}
\hat{\textbf{x}} \sim \mathcal{N}(\textbf{x}^{\star},\textbf{C}_{\hat{\textbf{x}}}),
\end{equation*}
where $\hat{\textbf{x}}$ is a $(d+1)$-dimensional vector,
$\textbf{C}_{\hat{\textbf{x}}}$ is a $(d+1)\times (d+1)$ matrix,
and~$\textbf{x}^{\star}$ is the true value of the unknown. For
example, when working in $\mathbb{R}^2$ their sizes become $3$ and
$3\times 3$, respectively.

\subsection{Taking the quadratic noise term~\eqref{eq:linprob} into
  consideration}
\label{sec:biased}
An alternative approach is to consider the linear model presented in
\eqref{eq:linprob}, where the noise term~$(\textbf{w}\circ\textbf{w})$
rules out Gaussianity, adding a bias to the expectation of the
estimator. Since this estimator is biased it does not guarantee that
minimum variance is achievable. The covariance matrix
$\textbf{C}_{\textbf{y}}$ for this kind of estimator will be slightly
different from~\eqref{eq:cyinvunb}.

The inverse of the covariance matrix of the measurements is now given
by
\begin{align}\label{eq:cyinv}
  \textbf{C}_{\textbf{y}}^{-1} = \begin{bmatrix}
    \frac{1}{4\|\textbf{s}_1-\textbf{p}\|^2\sigma_1^2+2\sigma_1^4} & & \multirow{2}{*}{\huge$0$}\\
    \multirow{2}{*}{\huge$0$} & \ddots &\\
    & & \frac{1}{4\|\textbf{s}_n-\textbf{p}\|^2\sigma_n^2+2\sigma_n^4}.
\end{bmatrix}
\end{align}
The estimator and its covariance matrix are again given by equations
\eqref{eq:estimator} and \eqref{eq:covest}, respectively, and the bias
is 
\begin{equation}\label{eq:biasf}
  E[\hat{\textbf{x}}]-\textbf{x} = (\textbf{A}^T\textbf{C}_{\textbf{y}}^{-1}\textbf{A})^{-1}\textbf{A}^T\textbf{C}_{\textbf{y}}^{-1}\bm{\Psi},
\end{equation}
where
\begin{gather}
\bm{\Psi} = \begin{bmatrix}
\sigma_1^2\\
\vdots \\
\sigma_n^2
\end{bmatrix}.
\end{gather}
The estimator results from a mixture between the Gaussian distribution
from the first noise term and a chi-squared distribution from the
second, with mean
$\textbf{x} +
(\textbf{A}^T\textbf{C}_{\textbf{y}}^{-1}\textbf{A})^{-1}\textbf{A}^T\textbf{C}_{\textbf{y}}^{-1}\bm{\Psi}$
and variance $\textbf{C}_{\hat{\textbf{x}}}$.

We see that the estimates will be biased, which  did not occur when
neglecting the squared noise term. In Section~\ref{sec:comp} we
numerically explore the trade-offs from choosing the unbiased,
twice-approximated estimator, and this approximation, encoded in~\eqref{eq:cyinv}.
Since the method in Section~\ref{sec:unbiased} uses a well
known closed-form estimator, we develop our distributed online active
localization algorithm based on the linear model~(\ref{eq:linunb}).

\section{Distributed ISEE.U estimator}
\label{sec:distr-isee.u-unbi}

Going back to the first method presented in
Section~\ref{sec:unbiased}, the linear estimator is a centralized
method, computed with full knowledge of all measurements and positions
of all agents, contained in $\textbf{y}$ and $\textbf{A}$,
respectively.  With the method developed in this section, each agent
will be able to compute an approximation to the linear estimator and
the covariance matrix for the estimator in a distributed way, using
only its own private data and data from one-hop neighbors.
\begin{remark}\label{thm:distr-observation}
  We now make a key observation: the estimator and the covariance
  matrix can both be computed using a sum of the contribution of every
  agent as
  \begin{equation}
    \hat{\mathbf{x}} = \Bigg(\sum_{i=1}^n \textbf{A}_i^TC_{\textbf{y}_{ii}}^{-1}\textbf{A}_i\Bigg)^{-1}\sum_{i=1}^n \textbf{A}_i^TC_{\textbf{y}_{ii}}^{-1}y_{i}
  \end{equation}
  \begin{equation}
    \textbf{C}_{\hat{\mathbf{x}}} = \Bigg(\sum_{i=1}^n
\textbf{A}_i^TC_{\textbf{y}_{ii}}^{-1}\textbf{A}_i\Bigg)^{-1},
  \end{equation}
  where each agent only has to provide its own line of matrix
  $\textbf{A}$, its own element of matrix
  $\textbf{C}_{\textbf{y}}$ and its own range measurement from
  $\textbf{y}$.
\end{remark}
Remark~\ref{thm:distr-observation} will be crucial in allowing for the
distributed greedy optimization of each agent's control.
Now, as in \cite{weng2014}, we will define matrix
$\bm{\mathcal{P}}_i(\tau)$ and vector $\textbf{z}_i(\tau)$ both
computed by agent $i$ at consensus iteration $\tau$. These are used to
compute the estimator and the covariance matrix for that agent. We
have
\begin{equation} \label{eq:xdist}
\hat{\textbf{x}}_i(\tau) = \bm{\mathcal{P}}_i^{-1}(\tau)\textbf{z}_i(\tau)
\end{equation}
\begin{equation} \label{eq:covdist}
\textbf{C}_{\hat{\textbf{x}}_i(\tau)} = \bm{\mathcal{P}}_i^{-1}(\tau).
\end{equation}
In contrast to~\cite{weng2014}, we state our novel consensus +
innovations iteration as
\begin{equation} \label{eq:psens}
\bm{\mathcal{P}}_i(\tau+1) = \frac{\tau}{\tau+1}\sum_{j \in N_i} \mathcal{W}_{ij}\bm{\mathcal{P}}_j(\tau) + \frac{1}{\tau+1}\sum_{j \in N_i} \textbf{A}_j^TC_{\textbf{y}_{jj}}^{-1}\textbf{A}_j
\end{equation}
\begin{equation} \label{eq:zsens}
\textbf{z}_i(\tau+1) = \frac{\tau}{\tau+1}\sum_{j \in N_i} \mathcal{W}_{ij}\textbf{z}_j(\tau) + \frac{1}{\tau+1} \sum_{j \in N_i} \textbf{A}_j^TC_{\textbf{y}_{jj}}^{-1}y_j(\tau+1)
\end{equation}
where $\bm{\mathcal{P}}_{i}$ is a $(d+1) \times (d+1)$ matrix and
$\textbf{z}_{i}$ is a $(d+1) \times 1$ vector, $N_i$ represents the
set of neighbors of agent $i$ (including $i$ itself), and where
$\bm{\mathcal{W}}$ is a weight matrix. Each agent knows only its own
line of the matrix, i.e., agent $i$ only knows $\bm{\mathcal{W}}_i$,
and this line contains the weights given by~$i$ to itself, and to its
neighbors on the communications graph $G(t)$. Matrix
$\bm{\mathcal{W}}$ is stochastic, i.e., its lines must add to
$1$. From now on we assume equal weights, despite the fact that our
solution can accommodate weight rebalance according to the information
reliability attributed to each neighbor.

In equation \eqref{eq:zsens} $\textbf{y}$ changes at every iteration
in the consensus phase: at every iteration every agent collects new
measurements that are introduced in the estimation process to reduce
the variance and improve accuracy. Introducing new data in the
consensus process is called consensus + innovations. A key novelty of
our iterations is to go beyond innovations for measurements collected
at $i$, and incorporate fresh measurements from neighboring
nodes. Similarly, the second term of the $\bm{\mathcal{P}}_i$
recursion is not the traditional consensus nor the traditional
consensus + innovations. This, as we will see, will give our method a
very clear advantage in mean error of the distributed quantities, for
finite time.
 
The work in~\cite{weng2014} presents a generic linear parameter
distributed estimation method that underperforms in our application,
as we will demonstrate with numerical results later on. We add
information of the neighbors in the second sum of equations
\eqref{eq:psens} and \eqref{eq:zsens}. In the initial iterations the
second sum has more relevance and effectively initializes both
quantities using the information that an agent can gather from its
neighbors and from itself. As $\tau$ increases the first sum becomes
dominant since both quantities already have the information from the
agent and its neighbors and we want to diffuse that information to the
rest of the network to reach a consensus estimator and covariance
matrix.

\subsection{Properties of ISEE.U at each node}
\label{sec:properties-isee.u-at}

We now establish that our estimator is centered, by stating that
ISEE.U asymptotically converges to the centralized unbiased estimator.
\begin{proposition}
  \label{th:1}
  For each node~$i$, the ISEE.U estimator~$\hat{\mathbf{x}}_i(\tau)$
  defined in~\eqref{eq:xdist} is centered, i.e.,
  \begin{equation}
    \label{eq:centered-expectation}
    \mathbb{E}_{\mathbf{x}}\left[ \hat{\mathbf{x}}_i(\tau) \right ] =
    \mathbf{x}, \text{ for all }\tau, \mathbf{x},
  \end{equation}
  where~$\mathbb{E}_{\mathbf{x}}\left[ X \right ]$ is the expected
  value of a random variable~$X$ with respect to the distribution
  of~$\mathbf{x}$.
\end{proposition}
A proof follows in Appendix~\ref{sec:proof-theorem}.  Our proposed
estimator is not efficient but simulations show that, for
finite consensus rounds, it achieves lower variance faster than other
efficient estimators, like the consensus + innovations. Further, it
has experimentally led to better estimates of uncertainty volumes
computed from covariance matrix estimates, as we will see later. In
conclusion, this section contains one of the major contributions of
this work since we presented a distributed method where, at each
iteration, agent~$i$ computes a centered estimate using
ISEE.U. Further analysis is required to rebalance the variance of the
estimator and better assess the finite and asymptotic behavior
of~$\tau \mathrm{Cov}(\hat{x}_i (\tau))$.

\subsection{Approximation quality of the ISEE.U estimator ---
  a numerical analysis}
\label{sec:ref}

To try to assess how accurate our localization estimates were, we
implemented a refinement phase after the target position
estimation. This way we could understand the impact of approximations
on our method in the estimation process. This phase takes the ISEE.U
position estimate and tries to improve it. To do so, we use
adaptations of two state-of-the-art methods: Distributed Gradient
Algorithm With \textit{Barzilai-Borwein} (BB)
Stepsizes~\cite{calafiore2012} and Distributed Maximum-Likelihood
network localization (GlobalSIP)~\cite{soares2014}.

Both methods are used for self localization of agents in a network
consisting of agents and anchors --- the only nodes that have full
knowledge of their positions. We applied these methods to perform the
self localization of the target using the agents of the network as
anchors. Both methods solve a minimization problem using the gradient
descent method~\cite{polyak} with spectral gradient (also known as \textit{Barzilai-Borwein}) stepsizes
\cite{fletcher2005,barzilai1988,birgin2000nonmonotone} and the projected gradient
method \cite{beck2013}, respectively, initialized with the
ISEE.U estimate.

The experiment consists in measuring the accuracy of the estimates of
the target position relative to the real one. To do so, the same setup
as in Section \ref{sec:comp} is used, where in each iteration the
randomly selected agent moves to the best position and through
consensus computes an estimate using ISEE.U. This estimate is then
used as initial estimate for the refinement methods. Finally, we
compute the
\textit{Mean
  Absolute Error} (MAE) given by
\begin{equation}\label{eq:MAEest}
\mathrm{MAE}(t) = \frac{\sum_{j = 1}^{M}{\|\textbf{p}(t)-\hat{\textbf{p}}_j(t)\|}}{M}
\end{equation}
for each of the three computed estimates. Here,~$\textbf{p}(t)$ is the
real target position at time $t$, $\hat{\textbf{p}}_j(t)$ is the
estimate of the target position in run $j$ at time $t$ and $M$ is the
number of Monte Carlo trials performed.
\begin{figure}[tb]
	\centering
		\includegraphics[width=0.6\linewidth]{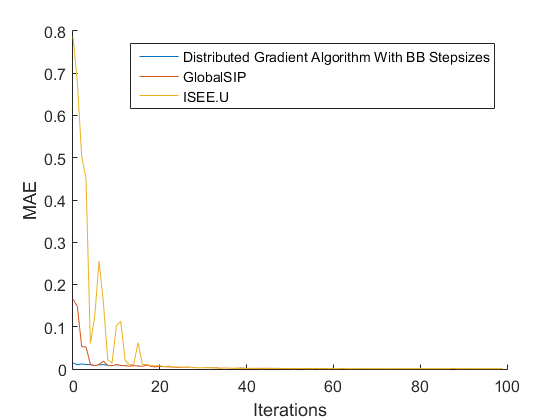}
                \caption[MAE for ISEE.U and for the two refinement
                methods.]{Mean Absolute Error (MAE) for ISEE.U and for
                  the two ML refinement methods initialized with the
                  ISEE.U estimate. The iterations axis describes the
                  overall estimation and control steps, so, as
                  iterations grow, the network improves the precision
                  of localization. The improvement of maximum
                  likelihood methods decays rapidly. After $20$ time
                  steps the accuracy of ISEE.U is no longer improved
                  by offline ML methods, which are computationally
                  heavier.}
\label{fig:rt1}
\end{figure}
The results of this experiment can be observed in
Figure~\ref{fig:rt1}. Here we can see that the refinement process
improves the estimation mostly in the first iteration, where the
agents are further away from the target and the measures have a larger
noise component. After approximately $20$ iterations, the network is
already close enough to the target to compute very accurate estimates
through the ISEE.U estimator, since the three methods produce almost
equal results. One can also notice that for some examples the
refinement even produces estimates with greater Mean Absolute Error
than ISEE.U: The Maximum-Likelihood cost function is parametrized by
the actual noisy range measurements, and the shape and maximum of the
likelihood function will change with different instances of these
measurements. Thus, the global minimum of the ML cost will not
coincide with the true positions for moderate noise levels and rigid
network configurations. For medium power noise we can even see the
localization error growing when the cost is declining.

Although the refinement process is very useful to improve our estimate
in certain cases, the two presented methods cannot be used directly
with our method, because they assume that the target is a cooperative
agent. The development of new refinement non-cooperative target
localization methods in a distributed setting is, thus, an important
avenue for future research.

\section{ISEE.Uctl: Nodes in motion}
\label{sec:isee.-nodes-moti}

For the control phase, termed ISEE.Uctl, we capitalize on
Remark~\ref{thm:distr-observation} and choose to move the network
so that the estimation accuracy of the target position is maximized,
within the reachable positions for each agent. We consider a random
deployment of the network's nodes and at each iteration each agent
will move to a reachable position where the localization error is
minimized. In order to do so we optimize a proxy for the localization
uncertainty and the optimization variable is sensor~$i$ position in the next time step,~$\mathbf{s}_{i}(t+1)$. Localization uncertainty will be measured as a
function of the covariance matrix of the estimator for each agent
$\textbf{C}_{\hat{\textbf{x}}_i}$ defined in \eqref{eq:covdist}, also
called precision matrix. Many authors take this same approach but
using the \textit{Fisher Information Matrix} (FIM), the inverse matrix
of the covariance.

We follow the \textit{D-optimality} \cite{boyd} criterion to solve the
minimization problem. Our method tries to minimize the volume of an
error ellipsoid, where each of its axes is obtained using the
eigenvalues from the covariance matrix. ISEE.Uctl tries to minimize
the length of all axes --- the same as minimizing all the eigenvalues
of the covariance matrix. To compute the cost function based on this
criterion we have to convert the value of the eigenvalues from the
covariance matrix into the length of the axes of the ellipsoid. This
is done by using
\begin{equation*}
l_j = \sqrt{\chi_{(d+1),\theta}^2\lambda_j}
\end{equation*}
where $l_j$ is half of the length of the axis associated with the
eigenvalue $\lambda_j$. The $\chi_{(d+1),\theta}^2$ is the value for
the chi-squared distribution with $d+1$ degrees of freedom and a
certain confidence level.

Then, the volume of the ellipsoid is given by
\begin{equation*} \label{eq:vols}
\begin{gathered}
V_{2\epsilon} = \frac{\pi^\epsilon}{\epsilon!}\prod_{j=1}^{2\epsilon} l_j\\
V_{2\epsilon+1} = \frac{2(\epsilon!)(4\pi)^\epsilon}{(2\epsilon+1)!}\prod_{j=1}^{2\epsilon+1} l_j
\end{gathered}
\end{equation*}
where $2\epsilon$ or $2\epsilon+1$ is the dimension of the
ellipsoid. This dimension should be equal to the dimension of the
covariance matrix and the number of eigenvalues.

Considering that the vector of solutions $\textbf{x}$ has always one
more dimension than the space we are working on, the ellipsoid
has that one extra dimension too. So, if our deployment area is in
$\mathbb{R}^2$, the ellipsoid will be in $\mathbb{R}^3$.

The procedure used by agent $i$ to compute its control is the
following:
\begin{enumerate}
\item Estimate the covariance matrix $\textbf{C}_{\hat{\textbf{x}}}$
  through ISEE.U;
\item Compute the volume for the current position of the network;
\item Subtract the contribution of agent~$i$ current position
  $(\textbf{A}_i^TC_{\textbf{y}_{ii}}^{-1}\textbf{A}_i)^{-1}$ from the
  covariance matrix according to
  Remark~\ref{thm:distr-observation},~$ \textbf{C}_{\hat{\mathbf{x}}}
  = \Bigg(\sum_{i=1}^n
  \textbf{A}_i^TC_{\textbf{y}_{ii}}^{-1}\textbf{A}_i\Bigg)^{-1},$ and
  save the resulting matrix;
\item Choose the new position;
\item Compute
  $(\textbf{A}_i^TC_{\textbf{y}_{ii}}^{-1}\textbf{A}_i)^{-1}$ for the
  new position;
\item Add its contribution to the resulting matrix from step $3$;
\item Compute the new volume;
\item Compare the volumes, and move to the position with smaller
  uncertainty volume.
\end{enumerate}

To make this solution realistic we discretize the deployment area, as
is usually done in the literature. When an agent wakes up, it follows
the procedure from step $4$ for all the eight possible positions (in
2D) around its current one. This is a greedy algorithm since each
agent makes the locally optimal choice every time it wants to move,
hoping that in the end the network can reach a global minimum.

In conclusion, we have now presented a way for the agents to move and
reconfigure the network, where the control for the movement is
computed considering the maximization of the estimation accuracy of
the target position.

\section{Numerical Results}
\label{cap:numres}

We show the performance of ISEE.U by comparing it with different
methods and under different scenarios. In Sections~\ref{sec:unbiased}
and \ref{sec:biased} we presented a linear estimator considering
Gaussian noise and a biased linear estimator, where a more faithful
mixture of Gaussian and~$\chi^{2}$ noise is taken into account. We
numerically compared ISEE.U and the biased linear estimator to
evaluate the impact of discarding the quadratic noise term, done in
ISEE.U. Then, we compared ISEE.U with asymptotically efficient
distributed estimators to demonstrate how our proposed estimator fares
against efficient state-of-the-art ones, and consider the trade-off in
asymptotic efficiency versus convergence speed.

We further refined the estimate of ISEE.U with two maximum likelihood
methods to numerically evaluate how far will our method be from the
true nonlinear maximum likelihood estimator.  Finally, the performance
of our algorithm was compared with a state-of-the-art method,
presented in~\cite{meyer2015}.  The error metric is the \textit{Mean
  Absolute Error} (MAE) defined in~\eqref{eq:MAEest}, and repeated
here for clarity
\begin{equation*}
\mathrm{MAE}(t) = \frac{\sum_{j = 1}^{M}{\|\textbf{p}(t)-\hat{\textbf{p}}_j(t)\|}}{M}.
\end{equation*}
Again, $\textbf{p}(t)$ is the real target position at time $t$,
$\hat{\textbf{p}}_j(t)$ is the estimate of the target position in run
$j$ and also at time $t$ and $M$ is the number of Monte Carlo trials
performed.

\subsection{Comparison between the unbiased and quadratic noise models}
\label{sec:comp}

We present a numerical comparison between the two methods to estimate
the target position: the ISEE.U estimator described in
Section~\ref{sec:unbiased}, and the biased linear estimator presented
in Section~\ref{sec:biased}, taking into account the quadratic noise
term in~\eqref{eq:linprob}.

\begin{figure}[tb]
	\centering
  		\includegraphics[width=0.5\linewidth]{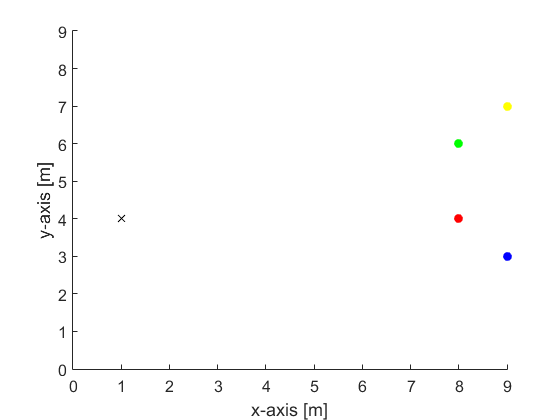}
                \caption{Initial setup. The circles represent the
                  agents, the black cross is the target
                  position. Agents start far away from the
                  target. Each agent will randomly start processing.
                  It will run the distributed ISEE.U estimator, and
                  the distributed control to not only localize the
                  target but also determine the agent's next best
                  position. An example of ISEE.U trajectory for a
                  static target can be seen at
                  \url{https://youtu.be/IrvHB5pHSxo}.} 
  \label{fig:iset}
\end{figure}
To test both algorithms we started with the setup presented in
Figure~\ref{fig:iset}. The space used is a $10 \times 10$ square and
the network has $4$ agents and they were initialized far away from a
static target so it was possible to see the error of the estimates for
different quantities of noise, as the variance of the noise depends on
the distance between the agents and the target. We ran~$100$
iterations of our algorithm where, in each iteration, one agent is
randomly chosen to move and the network enters the consensus +
innovations stage, running iterates~\eqref{eq:psens}
and~\eqref{eq:zsens}. Based on the estimated covariance matrix,
agent~$i$ chooses which is its next best position and moves to that
chosen position. The minimum number of iterations for the overall
network to move is $4$ and it can move up to $25$ times. This is more
than enough for the agents to reach their absolute best position given
the size of the space. We performed each run of $100$ iterations $20$
times and computed the MAE. The communication range of each agent was
defined as $55\%$ of the length of the diagonal of the square. This
means that if two agents are at a distance lower than $55\%$ of the
length of the diagonal of the square of each other they
communicate. The noise's standard deviation of each agent was
$\beta\hat{\sigma} = 0.001$ of the distance between the agent and the
target and the number of consensus iterations was defined as $T = 20$.

At each iteration a random agent chooses to move to the best position
and runs both ISEE.U and and the biased linear estimator presented in
Section~\ref{sec:biased} to compute an estimate for the target
position. Each estimate is then compared with the real position of
the target.
\begin{figure}[tb]
	\centering
		\includegraphics[width=0.6\linewidth]{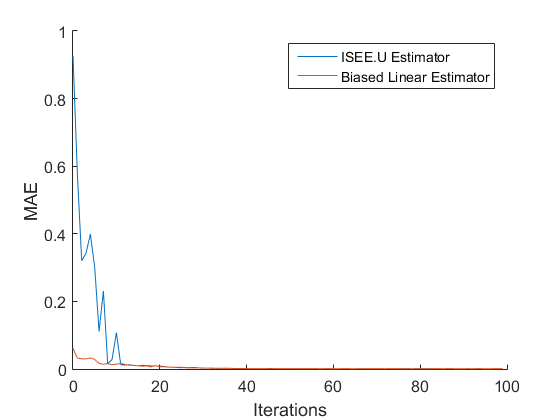}
                \caption{MAE for ISEE.U and for the estimator in
                  Section~\ref{sec:biased}, including quadratic noise
                  terms. The iterations axis represents the overall
                  estimation and control steps. The matched estimator
                  is more accurate than ISEE.U only in the first
                  iterations. After $15$ iterations their performance
                  is indistinguishable.}
\label{fig:MAEc}
\end{figure}
Figure~\ref{fig:MAEc} shows that around 10 iterations both estimators
reach approximately the same result, albeit the matched biased model
has a better transient. This is due to the mismatch between the data
noise and the noise model in~\eqref{eq:linunb}. However, the bias
in~\eqref{eq:biasf} was found to be small when compared to the error.
We could also observe that the movement of the agents is based on the
covariance matrix of the estimator. This covariance depends on the
covariance matrix of the measurements, given by~\ref{eq:cyinvunb} for
ISEE.U and by~\ref{eq:cyinv} for the biased linear estimator. Since
the covariance matrix of the measurements used in the biased linear
estimator is computed from the model without the extra approximation
it describes it better than the other matrix, so the control of the
agents with the biased linear estimator is better than the one with
ISEE.Uctl and this control can have influence on the position
estimates.

In conclusion, we can say that both estimators produce good estimates
for the target position with a slight advantage for the quadratic
noise estimator from the model in~(\ref{eq:linprob}). Also, the theory
backing up the convergence of the biased linear estimator is still to
be proven, which is a very promising avenue for future work.

\subsection{Comparison with other  consensus iterations}
\label{sec:distest}
An important contribution of this paper is a novel consensus +
innovations scheme that includes not only the node newly acquired
measurements, but also the recent measurements from one-hop
neighbors. How does this innovation fusion operates in terms of
convergence? And how does it impact not only the target position
estimate, but also the uncertainty volume used for control? These
questions led us to execute the following experiments.

We assessed the performance of our consensus + innovations against
efficient distributed consensus schemes for the linear
estimator. These methods are: (1) the basic consensus method, (2) the
traditional consensus + innovations algorithm and (3) a modified
consensus + innovations algorithm where the terms of the second sum in
equations \eqref{eq:psens} and \eqref{eq:zsens} were also multiplied
by the corresponding weights~$\mathcal{W}_{ij}$, as
\begin{equation*}
  \begin{split}
    \bm{\mathcal{P}}_i(\tau+1) =& \frac{\tau}{\tau+1}\sum_{j \in N_i}
    \mathcal{W}_{ij}\bm{\mathcal{P}}_j(\tau) \\
    &+ \frac{1}{\tau+1}\sum_{j
      \in N_i} \mathcal{W}_{ij} \textbf{A}_j^TC_{\textbf{y}_{jj}}^{-1}\textbf{A}_j \\
    \textbf{z}_i(\tau+1) =& \frac{\tau}{\tau+1}\sum_{j \in N_i}
    \mathcal{W}_{ij}\textbf{z}_j(\tau) \\
    & + \frac{1}{\tau+1} \sum_{j \in
      N_i} \mathcal{W}_{ij} \textbf{A}_j^TC_{\textbf{y}_{jj}}^{-1}y_j(\tau+1).
  \end{split}
\end{equation*}
\begin{figure}[tb]
	\centering
		\includegraphics[width=0.6\linewidth]{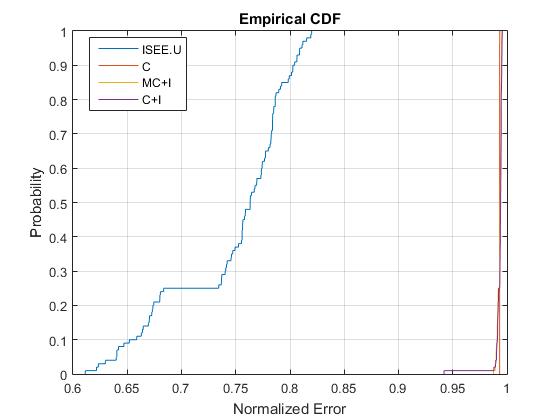}
                \caption[Empirical CDF of the normalized error for
                $\bm{\mathcal{P}}$ for the ISEE.U estimation part and
                other efficient estimators.]{Empirical CDF of the
                  normalized error of $\bm{\mathcal{P}}$ for the
                  ISEE.U estimator and multiple distributed efficient
                  estimators: consensus (C), consensus + innovations
                  (C+I) and modified consensus + innovations (MC+I)
                  for $20$ consensus rounds. ISEE.U outperforms the
                  other estimators in normalized error.}
	\label{fig:cdfp}
\end{figure}
One could expect that the above iteration would come to be as fast as
the ISEE.U iteration, because they also include extra new measurements
from neighbors. Nevertheless, our experiments deny this hypothesis, as
will be seen in Figure~\ref{fig:cdfp}.  For this experiment we
considered a network of $100$ agents randomly deployed in a
$100 \times 100$ square. The communication range was lowered, defined
as $30\%$ of the length of the diagonal of the square, the noise's
standard deviation of each agent was again $\beta\hat{\sigma} = 0.001$
of the distance. We ran the experiment and computed the error
between~$\bm{\mathcal{P}}$ obtained for each method and the ML one for
each agent of the network, given by the Frobenius norm of the error
matrix~$\|\bm{\mathcal{P}}_{\mathrm{method},i}-\bm{\mathcal{P}}_{ML}\|_{F}$
and computed the empirical \textit{Cumulative Distribution Function}
(CDF). To better understand the results the error was normalized with
the norm of $\bm{\mathcal{P}}_{ML}$. An equivalent test was done for
$\textbf{z}$.
\begin{figure}[tb]
	\centering
		\includegraphics[width=0.6\linewidth]{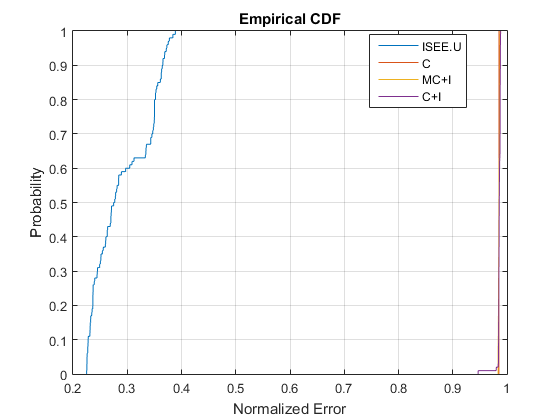}
                \caption[Empirical CDF of the normalized error for
                $\textbf{z}$ for ISEE.U and multiple efficient
                estimators.]{Empirical CDF of the normalized error of
                  $\textbf{z}$ for the ISEE.U estimator and multiple
                  distributed efficient estimators: consensus (C),
                  consensus + innovations (C+I) and modified consensus
                  + innovations (MC+I) for $20$ consensus
                  rounds. ISEE.U outperforms the other estimators.}
	\label{fig:cdfz}
\end{figure}
From Figures~\ref{fig:cdfp} and~\ref{fig:cdfz} we conclude that ISEE.U
clearly outperforms the other distributed estimators from observing
that empirical probability of error is much smaller for our novel
consensus + innovations scheme, with the iterates in~\eqref{eq:psens}
and~\eqref{eq:zsens}, thus supporting the claims made in
Section~\ref{sec:method} where we stated that ISEE.U had a notorious
advantage in mean error.

\begin{figure}[tb]
	\centering
		\includegraphics[width=0.6\linewidth]{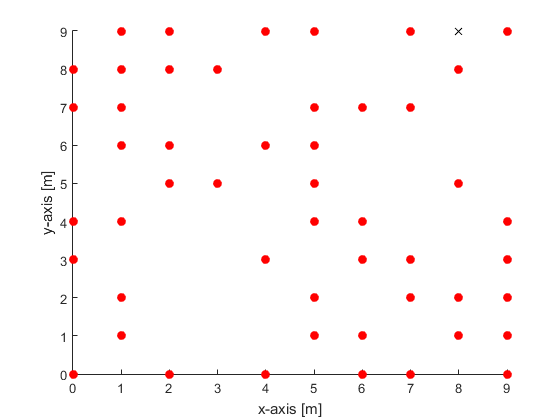}
	\caption{Setup used to compute the variance of the estimator:
          a geometric network of~50 agents randomly deployed on a
          grid (red dots), and one target (black cross).}
	\label{fig:setupvar}
\end{figure}

To further validate ISEE.U we also evaluated the behavior of the
variance of the estimator with the number~$\tau$ of consensus
iterations. To do so we considered a geometric network of $50$
randomly deployed agents and one target presented in
Figure~\ref{fig:setupvar}. The communications range of each sensor was
manipulated so that the average node degree\footnote{To characterize
  the network we use the concepts of node degree~$k_{i}$, the number
  of edges linked to node~$i$, and average node degree
  $\langle k \rangle = 1/n \sum_{i=1}^{n} k_{i}$.} was $8.44$. The
noise standard deviation for each agent measurements was again defined
as $\sigma_i = 0.001\|\textbf{s}_i - \textbf{p}\|$. We ran $1000$
Monte Carlo trials so that we could estimate the covariance of
$\textbf{z}$ and the variance of the estimator for one agent, defined
as~$\mathrm{var}(\hat{\textbf{x}}(\tau)) = \textbf{tr}
(\text{Cov}(\hat{\textbf{x}}(\tau)))$, where~$\textbf{tr}(\cdot)$ is
the trace linear operator.
\begin{figure}[tb]
\centering
\begin{subfigure}{0.45\textwidth}
  \centering
  \includegraphics[width=1\linewidth]{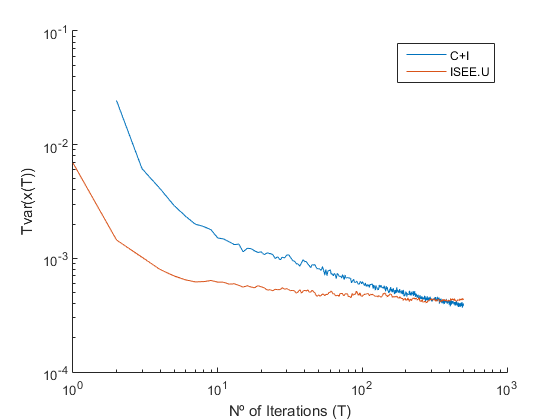}
  \caption{}
  \label{fig:covest}
\end{subfigure}
\begin{subfigure}{.45\textwidth}
  \centering
  \includegraphics[width=1\linewidth]{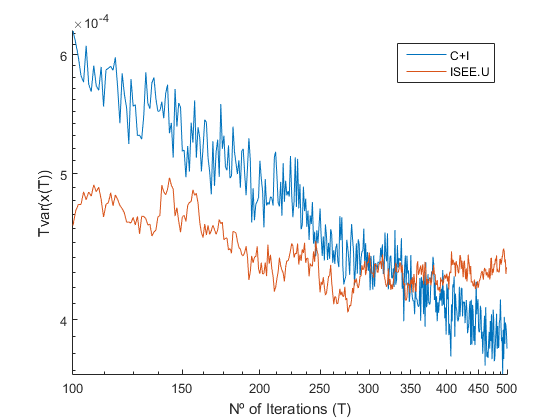}
  \caption{}
  \label{fig:covzoom}
\end{subfigure}
\caption[Variance of the estimator for ISEE.U and consensus +
innovations (C+I).]{Variance of the estimator for ISEE.U and consensus
  + innovations (C+I). ISEE.U is very fast reaching a plateau in
  approximately $10$ consensus iterations being surpassed by C+I after
  approximately $350$ iterations. This represents more than one order of
  magnitude less iterations. \textbf{(b)} is a detail of
  \textbf{(a)}. In \textbf{(a)} the C+I has no value for the first
  iteration because matrix~$\bm{\mathcal{P}}_{i}(1)$ only takes~$i$ own
  measurements, and so it is not invertible.}
\label{fig:covarest}
\end{figure}
Figures~\ref{fig:covest} and~\ref{fig:covzoom} show that ISEE.U is
faster than the traditional consensus + innovations method in
attaining a low variance.  ISEE.U reaches a plateau of low variance in
roughly $10$ consensus iterations while the traditional consensus
needs approximately $350$: more than one order of magnitude more
iterations. Since the variance goes to $0$ because more data is
introduced to the estimator during consensus we need to evaluate the
product of $\tau \: \mathrm{var}(\textbf{x}_i(\tau))$ instead of only
$\mathrm{var}(\textbf{x}_i(\tau))$ to be able to visualize the plateau
of minimum variance. As expected, ISEE.U is not efficient, i.e., its
variance does not touch the CRLB, unlike the vanilla consensus +
innovations. Thus, the plateau
of~$\tau \mathrm{var}(\textbf{x}_i(\tau))$ for consensus + innovations
is indeed smaller, as depicted in
Figure~\ref{fig:covarest}. Figure~\ref{fig:covzoom} evidences that the
trade-off is very small.  Although not attaining the CRLB, ISEE.U
decreases estimation variance to reasonable values with one order of
magnitude less iterations.

\subsection{Simulations with a state-of-the-art method}
\label{sec:sims}

The experiments provided in this section assess the relevance of the
proposed method of active target localization, when compared with the
state-of-the-art. As discussed in Section~\ref{sec:related-work}, the
algorithm that better matches our problem formulation
is~\cite{meyer2015}, with which we will numerically compare in terms
of performance.

We start by defining different parameters that were used in the
simulation. The space considered is a $200 \times 200$ square and,
thus, our grid now has $40000$ possible positions for the agents. The
standard deviation for the noise continues to depend on the distance
and $\beta\hat{\sigma} = 0.1$ and was applied to both methods.  For
ISEE.U we maintained a communication range of $55\%$ of the length of
the diagonal of the space for each agent. For fairness, the motion of
the network in ISEE.U was altered. Previously, each agent decided to
move randomly. Now, at each iteration, agents operate sequentially, so
that in every iteration all agents have the opportunity to move inside
their possible motion range. After each agent's movement, the network
computes an estimate for the target position through consensus.

For the benchmark and following the Simulations section
in~\cite{meyer2015},~$120,000$ samples were used in the estimation
layer and~$6,000$ samples in the control layer.

For a defined number of iterations we ran both algorithms, and each
produced an estimate for the target position and the control. This
process was repeated $5$ times so that the target estimate from both
methods could be used to compute the MAE of estimation error.

The simulation consisted in the usual network of four agents and one
mobile target. The agents were deployed in random positions and the
simulation ran for $150$ iterations. At each run the target state at
some iteration~$t$,
$\textbf{x}_{\mathrm{target}}(t) = [p_1(t) \: \: p_2(t) \: \:
\dot{p}_1(t) \: \: \dot{p}_2(t)]^T$ changes according to
\begin{equation*}
\textbf{x}_{\mathrm{target}}(t) = \textbf{x}_{\mathrm{target}}(t-1) + \bm{\Gamma}(t) + \Lambda \textbf{q}(t),
\end{equation*}
where $\bm{\Gamma}(t)$ is
\begin{equation*}
\begin{gathered}
\bm{\Gamma}(t) = \begin{bmatrix}
R_t(t)\cos(\pi-\frac{v}{R(t)}t)+ c_1 \\
R_t(t)\sin(\pi-\frac{v}{R}t)+ c_2 \\
0 \\
0
\end{bmatrix}, 
\end{gathered}
\end{equation*}
$\textbf{q}(t) \in \mathbb{R}^2$ where
$q_i(t) \sim \mathcal{N}(0,\tilde{\sigma}_q^2)$ is Gaussian with zero
mean and variance $\tilde{\sigma}_q^2 = 10^{-5}$ and where $v$ is the
linear velocity of the target that was kept constant and defined as
$70\%$ of the velocity of the group, $R_t(t)$ is the radius of the
spiral that was decreased every $10$ iterations according to
$R_t(t) = 0.97R_t(t-1)$, where $R_t(0) = 20$, and
$\textbf{c} = [c_1 \: \: c_2]^T$ are constants defining the center of
the spiral, and chosen as $\textbf{c} = [20 \: \: 0]^T$. So the target
will start at its normal starting position and will move clockwise in
a spiral trajectory centered in $[70 \; \: 0]^{T}$.
\begin{figure}[tb]
\centering
\begin{subfigure}{0.45\textwidth}
  \centering
  \includegraphics[width=1\linewidth]{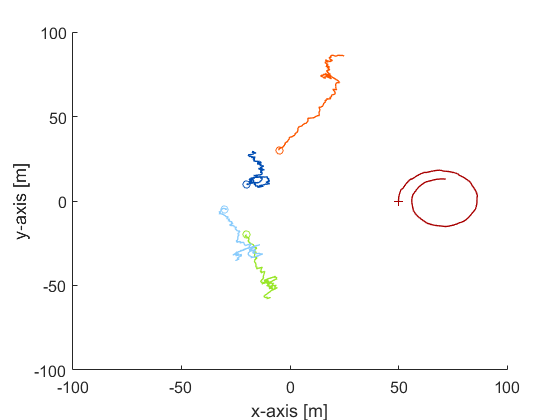}
  \caption{Benchmark. The benchmark method cannot follow a target
    describing a mismatched motion model. See sample trajectory at
    \url{https://youtu.be/kNium47NpGg}.}
  \label{fig:bench_1e4}
\end{subfigure}
\begin{subfigure}{.45\textwidth}
  \centering
  \includegraphics[width=1\linewidth]{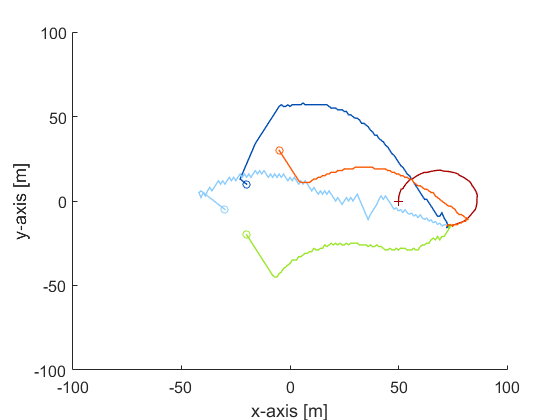}
  \caption{ISEE.U. Our proposed method is agnostic to the type of
    target motion. See sample trajectory at \url{https://youtu.be/LqMLo1AGZQY}.}
  \label{fig:mvu_1e4}
\end{subfigure}
\caption[Trajectories for both methods.]{Sample trajectories for both
  methods. The trajectories are represented by lines having the same
  color of the circle of the agent that they correspond. The red line
  represents the trajectory of the target. The agents in \textbf{(a)}
  spread and loose track of the target, while in \textbf{(b)}
  the agents pursue and trap the target.}
\label{fig:sim_1e4}
\end{figure}
Figures~\ref{fig:sim_1e4} illustrate differences between trajectories
of both methods.  For the benchmark, we can see in
Figure~\ref{fig:bench_1e4} that the agents start approaching the
target and when they reach a distance from which it is not possible to
decrease measurement noise, the agents spread out to find a formation
more suitable to locate and track the target. The belief should also
be represented in the form of samples, but in the final moments of the
pursuit task, this belief is very far away from the actual target
position: the benchmark formulation assumes a linear trajectory for
the target. When the target changes its motion, agents still believe
that the target is moving in a straight line, and continue to produce
consistent estimates with this model until reaching the boundary of
the space. In fact, agents start to move in the direction of the
target, but since they start computing wrong estimates, they spread to
try to minimize that error.

With ISEE.U, as seen in Figure~\ref{fig:mvu_1e4}, the agents move in
the direction of the target and inexorably end up trapping it. In fact,
the cost function used to compute the control is the volume of the
error ellipsoid, which is a function of the covariance matrix of the
estimator from equation~\eqref{eq:covest}. Thus, it depends on the
distance between the agents and the target via the covariance of the
measurement data represented in equation \eqref{eq:cyinvunb}. As the
error is dependent on the distance, agents try to minimize their
distance to the target.
\begin{figure}[tb]
	\centering
		\includegraphics[width=0.6\linewidth]{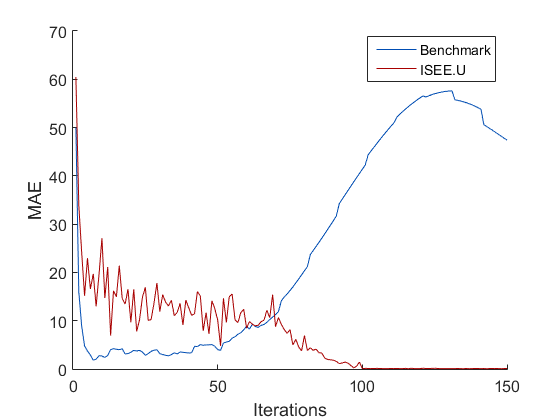}
                \caption{MAE obtained for both methods in the
                  simulation. ISEE.U outperforms the benchmark after
                  approximately $60$ iterations. After $50$ iterations
                  the MAE for the benchmark increases, documenting the
                  fact that the benchmark algorithm no longer tracks
                  the target, while ISEE.U, assuming no model for the
                  target motion, can equally compensate the spiral
                  movement.}
	\label{fig:mae_1e4}
\end{figure}
In Figure~\ref{fig:mae_1e4} we can see that, after approximately~$10$
iterations, the benchmark reaches a low MAE and stays around the same
value until approximately the $50$-th iteration, where the MAE starts
increasing. This coincides with the moment when the movement of the
target changes and so the estimates start to be further away from the
real target position. Thus, the estimation error increases. ISEE.U
needs more iterations to reach a low MAE but, after after
approximately~$60$ iterations, it reaches the same value as the
benchmark. As the network of ISEE.U agents has no assumption on the
target movement, ISEE.U continues decreasing its error until reaching
a MAE very close to $0$, outperforming the benchmark in this
last stretch.

\section{Conclusions and future work}
\label{sec:conclusion}


In this paper we addressed active estimation of target positions using
a network of mobile agents in a distributed way. Our proposed method,
ISEE.U, is in general more accurate than the benchmark when no
specific dynamical model is assumed for the target. Our ISEE.U overall
scheme of estimation and control is simple to implement, runs
distributedly --- and even asynchronously, with good numerical
performance, but also with the guarantee that, at each node, the
expected value of the position estimate is equal to the centralized
maximum-likelihood estimator.

As a byproduct, we developed a novel consensus +
innovations scheme that numerically achieves more precision that the
known consensus and consensus + innovations algorithms, for the same
number of consensus rounds.
In fact, our ISEE.U estimator, despite the variance plateau being a
bit higher than the Cram\'{e}r-Rao bound, could achieve, in our
simulations, more precision than the efficient traditional consensus
and consensus + innovations for the same number of consensus rounds.

The control takes advantage of two key facts that have gone unnoticed
until now: (i) in our distributed scheme every agent has access to an
estimate of the covariance matrix for the overall estimator, (ii) the
overall covariance can be seen as a sum of each agent's
contributions. These two facts imply that an agent can take its
estimate of the covariance for the whole network and test which future
movement will minimize the uncertainty in the estimation.

When comparing the ISEE.U active positioning algorithm with a
state-of-the-art solution under different scenarios, numerical results
indicate that, for a static or linear motion target, the benchmark
increases localization precision at a faster pace in the beginning of
the trajectory, but that ISEE.U catches up and even increases
localization precision when compared with the benchmark. For more
varied trajectories the results are even more interesting: ISEE.U
clearly outperforms the benchmark in localization precision. The
running times are also very competitive: ISEE.U takes a few seconds,
being capable of online operation,
while the benchmark --- a solution based on a particle filter --- can
run for hours, thus not adapted to real-time deployment.

In conclusion, we designed a flexible method that is easy to implement
and to adapt to very different scenarios and that, on top of this,
competes in terms of performance with a much more complex
state-of-the-art method.

\section*{Acknowledgment}
The authors would like to thank Prof.\ F. Meyer for providing the implementation of his published
algorithm.


\appendix

\section{Proof of Proposition~\ref{th:1}}
\label{sec:proof-theorem}

We start by computing the expected value of $\textbf{z}_i(\tau+1)$, given by equation \eqref{eq:zsens}, resulting in
\begin{equation*}
  \begin{split}
    \mathbb{E}_\textbf{x}[\textbf{z}_i(\tau+1)] &= 
    \frac{\tau}{\tau+1}\sum_{j \in \mathcal{N}_i}
    \mathcal{W}_{ij}\mathbb{E}_\textbf{x}[\textbf{z}_j(\tau)] \\
    &+
    \frac{1}{\tau+1} \sum_{j \in \mathcal{N}_i}
    \textbf{A}_j^TC_{\textbf{y}_{jj}}^{-1}\mathbb{E}_\textbf{x}[y_j(\tau+1)],
  \end{split}
\end{equation*}
since $\mathbb{E}_\textbf{x}[\textbf{y}(\tau+1)] = \textbf{A}\textbf{x}$ and considering $\bm{\mathcal{P}}_i(\tau+1)$, given by equation \eqref{eq:psens}, we get
\begin{equation*}
E_\textbf{x}[\textbf{z}_i(\tau+1)] = \bm{\mathcal{P}}_i(\tau+1)\textbf{x}.
\end{equation*}
Considering equation \eqref{eq:xdist}
\begin{equation*}
\begin{split}
\mathbb{E}_\textbf{x}[\hat{\textbf{x}}_i(\tau)] &= \bm{\mathcal{P}}_i^{-1}(\tau)\mathbb{E}_\textbf{x}[\textbf{z}_i(\tau)] \Leftrightarrow  \\
 &= \bm{\mathcal{P}}_i^{-1}(\tau)\bm{\mathcal{P}}_i(\tau)\textbf{x} \Leftrightarrow \\
 &= \textbf{x} \: \: \forall_{\tau,\textbf{x}}. 
\end{split}
\end{equation*}
This proves that the ISEE.U distributed estimator is centered.

\phantomsection
\addcontentsline{toc}{chapter}{Bibliography}
\bibliographystyle{elsarticle-num}
\bibliography{02.biblio}

\end{document}

%% file: 03.abstract.tex
This paper addresses target localization with an online active learning algorithm defined by distributed, simple and fast computations at each node, with no parameters to tune and where the estimate of the target position at each agent is asymptotically equal in expectation to the centralized maximum-likelihood estimator. 
ISEE.U takes noisy distances at each agent and finds a control that maximizes localization accuracy. We do not assume specific target dynamics and, thus, our method is robust when facing unpredictable targets. Each agent computes the control that maximizes overall target position accuracy via a local estimate of the Fisher Information Matrix.  We compared the proposed method with a state of the art algorithm outperforming it when the target movements do not follow a prescribed trajectory, with x100 less computation time, even when our method is running in one central CPU.